\documentclass[journal]{IEEEtran}
\usepackage{amsmath,amsfonts}
\usepackage{algorithmic}
\usepackage{array}
\usepackage[caption=false,font=normalsize,labelfont=sf,textfont=sf]{subfig}
\usepackage{textcomp}
\usepackage{stfloats}
\usepackage{url}
\usepackage{verbatim}
\usepackage{graphicx}
\usepackage{amssymb}  
\usepackage{footnote}
\usepackage{xcolor} 
\usepackage[hidelinks]{hyperref}    
\usepackage{lipsum} 
\usepackage{multirow}
\usepackage{multicol}
\usepackage{bm}
\usepackage{booktabs,caption}
\usepackage[flushleft]{threeparttable}
\usepackage{adjustbox}

\def\BibTeX{{\rm B\kern-.05em{\sc i\kern-.025em b}\kern-.08em
    T\kern-.1667em\lower.7ex\hbox{E}\kern-.125emX}}
\usepackage{balance}
\begin{document}
\title{A Dual-Motor Actuator for Ceiling Robots with High Force and High Speed Capabilities}
\author{Ian Lalonde$^{1}$, Jeff Denis$^{1}$, Mathieu Lamy$^{1}$ and Alexandre Girard$^{1}$
\thanks{This work was supported by \textcolor{red}{the Fonds Québécois de la recherche sur la nature et les technologies (FRQNT) and the Natural Sciences and Engineering Research Council of Canada (NSERC).}}
\thanks{$^{1}$All authors are with the Department of Mechanical Engineering, Université de Sherbrooke, QC, Canada.}%
}


\maketitle

\begin{abstract}
Patient transfer devices allow to move patients passively in hospitals and care centers. Instead of hoisting the patient, it would be beneficial in some cases to assist their movement, enabling them to move by themselves and reducing hospitalization time. 
However, patient assistance requires devices capable of precisely controlling output forces at significantly higher speeds than those used for patient transfers alone, and a single motor solution would be over-sized and show poor efficiency to do both functions. 
This paper presents a ceiling robot, using a dual-motor actuator and adapted control schemes, that can be used to transfer patients, assist patients in their movement, and help prevent falls.
The prototype is shown to be able to lift patients weighing up to 318~kg, and to assist a patient with a desired force of up to 100~kg with a precision of 7.8\%. 
Also, a smart control scheme to manage falls is shown to be able to stop a patient who is falling by applying a desired deceleration.
\end{abstract}


\section{INTRODUCTION}

Patient transfer devices are commonly used in medical facilities to help caregivers move patients at a controlled pace by supporting their full weight. 
Studies suggest self-movements for patients with the device assisting them are beneficial \cite{SUMIDA2001391} \cite{Dirkes} instead of lifting their full weight. This can help improve healing time and reduce rehospitalization risks~\cite{Functionalstatus}\cite{Motorandcognitive}.
While assistive equipment \cite{Safegait360} \cite{ErgoTrainer} \cite{aretech} have been successfully utilized in rehabilitation settings, their widespread implementation in hospital rooms remains cost-prohibitive.
Improving existing patient transfer equipment \cite{Handicare} \cite{ms2} \cite{Savaria} with rehabilitation features can enhance patient mobility. This approach ensures continued support for patients with limited mobility while facilitating rehabilitation exercises for those capable of more independent movements.

Commercial lifting devices aim for versatility for both patient transfer and patient assistance \cite{guldmann}, yet their maximum speed of 0.1 m/s falls short for certain daily activities. For instance, during a sit-to-stand (standing up from a sitting position), the center of mass of a patient can reach a vertical velocity of 0.35~m/s \cite{ROEBROECK1994235}, making this system unable to follow the patient for all their movements. 
Using a smaller gear ratio for the actuator can increase it's maximum speed and enable following the patient during rehabilitation. Unfortunately, this decreases the maximum force at the output making this solution unsuited for both applications.

Therefore, the mechanical requirements for patient assistance differ significantly from those for patient transfer. 
For patient transfer, the lift winches the full patient's weight (high force) and uses velocity control to move the patient.
For patient assistance, the lift unloads a certain percentage of the patient's weight while following their movements. This means working at high speed and low force, and controlling the force applied on the user.
The lift also needs to be able to prevent the patient from falling to the floor (fall prevention).

\begin{figure}[!t]
    \centering
    \includegraphics[width=\linewidth]{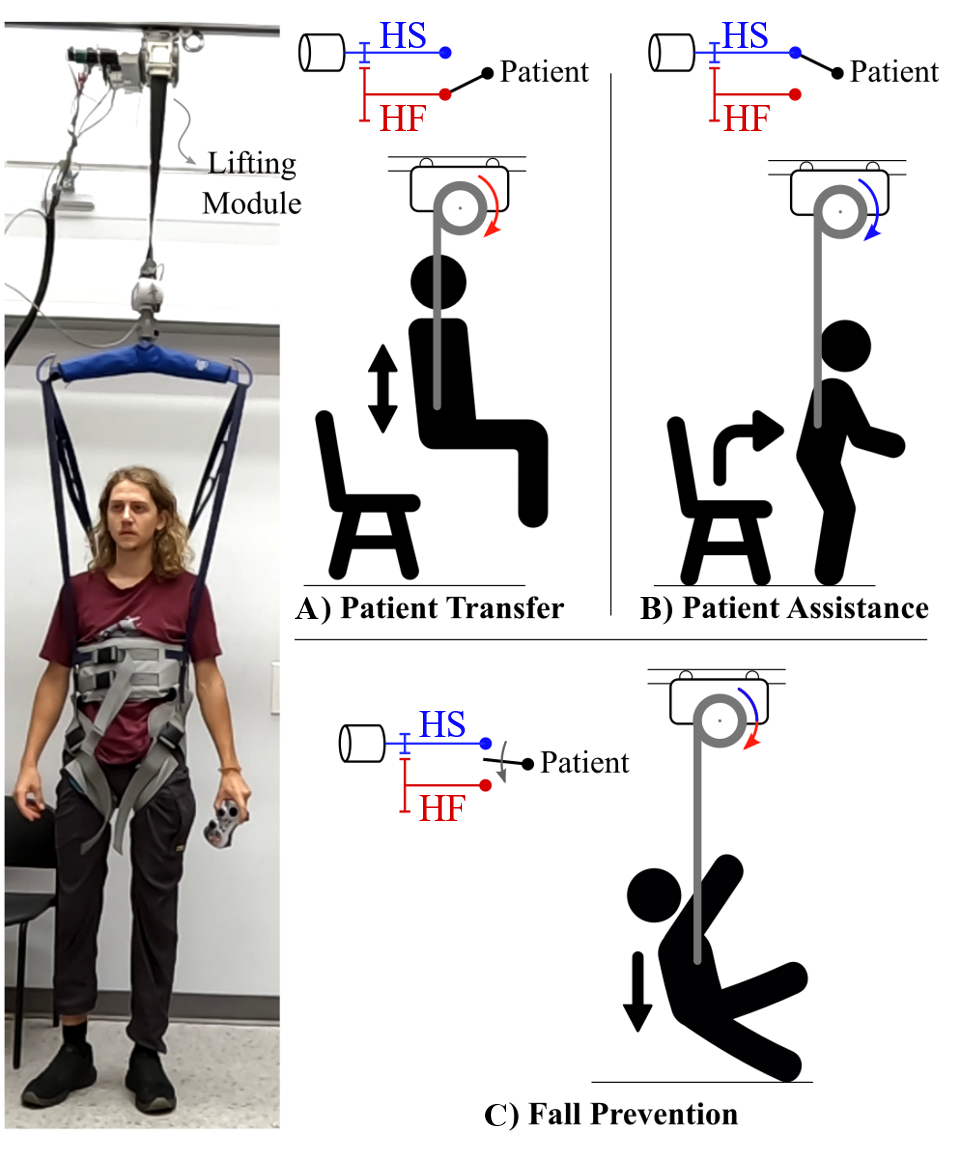}
    \caption{Lifting module overview of the prototype in a real situation with illustrations of its three operating conditions, A) Patient Transfer, B) Patient Assistance, C) Fall prevention. HF is for highly-geared mode (high-force) and HS is for lightly-geared mode (high-speed).}
    \label{fig:system_overview}
\end{figure}

Addressing the divergent requirements of patient transfer and assistance with a single-motor actuator presents challenges due to the trade-offs between force and speed capabilities. While lightly-geared electric motors excel in low-force, high-speed applications, they prove inefficient for high-force tasks typical in patient transfer scenarios. Increasing the gear ratio like in available transfer devices limits the maximum velocity and increases the reflected inertia and friction, penalizing the force tracking accuracy. 

Alternative approaches have been explored to tackle similar problems:
\textbf{1) Oversized electric motor:} 
One simple solution to reach both modes of operation is to use a bigger motor with enough force and speed capabilities for patient transfer and rehabilitation \cite{aretech} \cite{Frey}. This solution increases the actuator's weight (and therefore costs). For a ceiling base actuator, this also means the patient needs to carry a heavier device while walking during rehabilitation. This is undesirable for patients who already need assistance while walking.  
%
\textbf{2) Force feedback:}
To control the output, a force sensor can be mounted in series with the load and use its signal for a closed-loop force control algorithm \cite{aretech} \cite{Rad} \cite{DOB_BWS}. A friction compensation algorithm can also be used instead to control the force without the need for a load cell \cite{GUVENC1994623}. Despite using a force control algorithm, the performances are limited by the actuator used.  
\textbf{3) Series elastic actuator (SEA):}
SEA have been explored in research for patient assistance devices to improve force control by negating motor friction and inertia \cite{aretech} \cite{MacLean} \cite{lokolift}, but the spring limits the displacement for the assistance mode and needs to be locked before transfer mode to support the full weight of the patient. This is problematic for care centers since it adds time for patient transfer tasks. Adding a spring also doesn't improve the maximum speed during rehabilitation which is again limited by the actuator. 
%
%
\textbf{4) Variable gear-ratio systems:} Similarly to car powertrains, variable gear-ratios could address the divergent requirements for both applications without oversizing the motor. The system could downshift to a large reduction ratio to achieve high force capabilities during patient transfer and upshift to a small reduction ratio for high-speed capabilities while assisting a patient.
However, transitioning from high-speed to high-force is crucial to safely stop a patient from falling during a training session, which is not feasible with a regular variable gear-ratio system \cite{Girard2015}.  

A dual-motor actuator can offer seamless shifting between two distinct configurations since both motors can be connected to the output at the same time. This type of actuator has been explored for a wide range of robotic applications. 
T.~Takayama and al. \cite{Takayama} have used a force-magnification drive for a robotic hand that can move at high speed and apply high force when gripping an object. But the high-speed mode is not backdrivable which would penalize force tracking accuracy if used with a force controller. 
To overcome those problems, 
A.~Lecavalier and al. \cite{lecavalier2022bimodal} present a bimodal hydrostatic actuator for robotic legs with one strong mode when the leg is supporting the weight and one fast mode for the foot placement when the leg is swinging freely. This solution leads to a lighter and more efficient actuator \cite{Denis2022}, but the range of motion is limited by the stroke of the hydraulic cylinder. 
A.~Girard and al. \cite{Girard2015} have developed a dual-motor actuator with a planetary differential gearbox and locking brakes to control which motor is giving energy to the output. This offers a compact solution with an unlimited range of motion at the output \cite{VERSTRATEN2018134}. 
However, the transition between both modes needs to happen at low speed due to the brake's limitations. This is problematic for fall prevention since the system needs to take a high load at high speed. Having a brake with controllable force output, like a disk brake, would help control the deceleration of the patient and improve their comfort.  In summary, although dual-motor actuators show promise for applications with discrete points of operation, applying this technology to patient handling equipment is not explored yet and includes specific technical challenges such as managing smooth transitions while fighting a large backdriving force for fall prevention. Addressing those technical challenges with a modified mechanical design and control scheme is the main contribution of this paper.

This paper presents the design and control of a novel dual-motor multifunctional ceiling-based device that extends the capabilities of current healthcare lifting devices up to the capabilities of current assistive devices, with little compromise on system mass and efficiency. 
Figure~\ref{fig:system_overview} gives an overview of the system and its operating modes. 
The contributions are: a) a novel control scheme for a dual-motor actuator with a disk brake for a seamless transition at high speed and high force (a specific requirement for fall prevention), b) a comparative analysis of four actuator design architectures, c) a novel friction compensation algorithm that leverage the nullspace of the dual-motor actuator, and d) a full-sized, fully functional prototype of a multi-function ceiling lift using the proposed dual-motor actuator technology. 
Section~\ref{sec:requirements} presents the system requirements. In section~\ref{sec:actuator}, the selected motor architecture is introduced and compared to other relevant designs. This section also describes the dynamical model and the research prototype designed. Section~\ref{sec:algo} presents control algorithms for three operating modes: patient transfer, patient assistance, and fall prevention with experimental results shown in section~\ref{sec:exp}. 

\section{System Requirements} \label{sec:requirements}
\begin{figure}[h]
    \centering
    \includegraphics[width=0.7\linewidth]{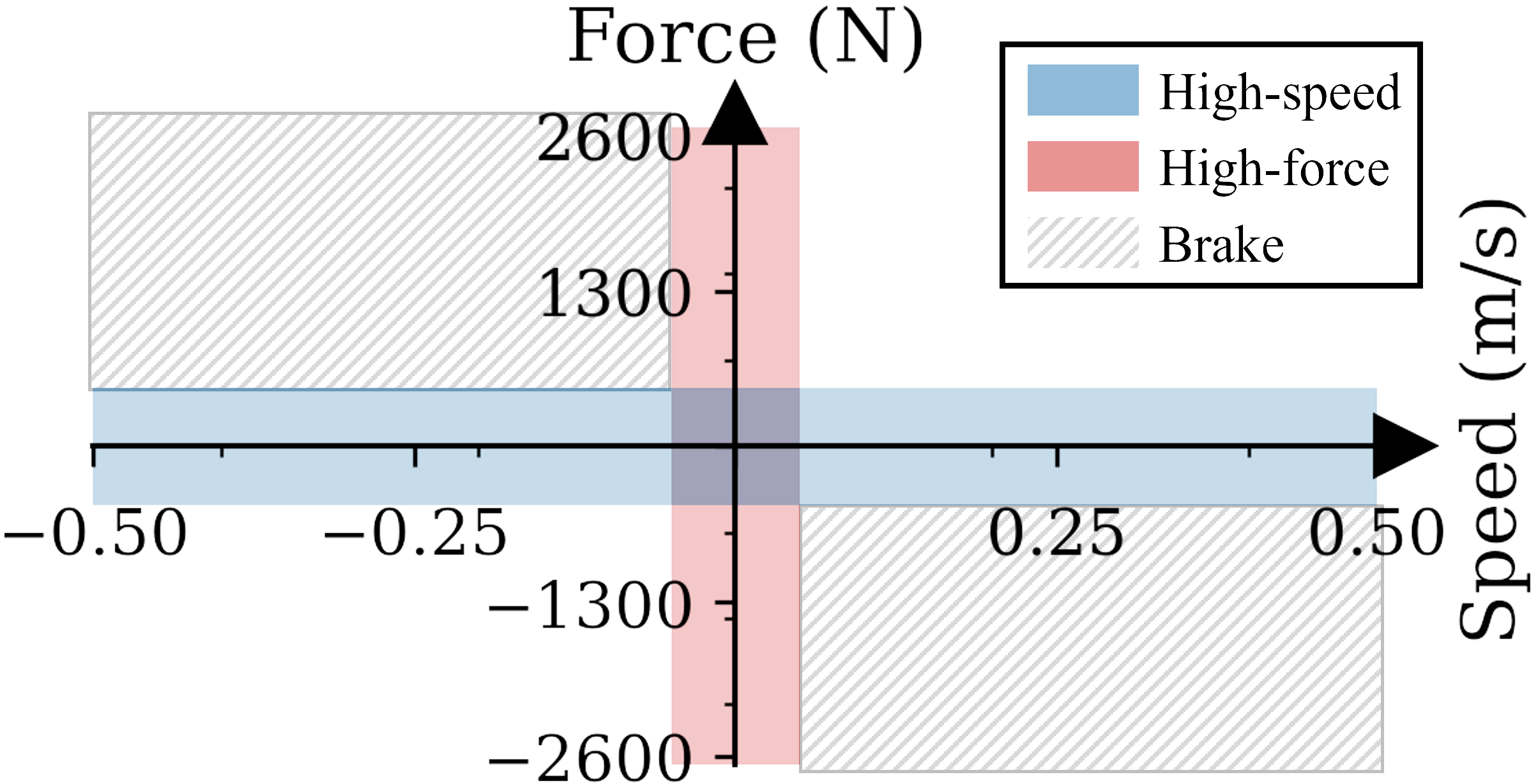}
    \caption{Operation points mapping (continuous force) with high-force for patient transfer, high-speed for patient assistance and brake for fall prevention.}
    \label{fig:point_operation}
\end{figure}

The multifunctional lift is split into three operating conditions. Figure~\ref{fig:point_operation} maps these operating conditions on a force-velocity plot. This section describes the target requirements.

\subsection{Patient Transfer}
Patient transfer is for hoisting patients up to 272~kg (600 lb) at 0.04 m/s (107~W). This fits within the range of typical medical lifting devices, which can lift patients weighing between 250 and 450~kg at a speed between 0.03 and 0.05~m/s \cite{Handicare} \cite{ms2} \cite{Savaria}. Safe and precise manipulation of the patient limits the desired speed in this mode.

\subsection{Patient Assistance}
Patient assistance is for assisting the muscles of patients during their daily activities. It is set to 50~kgf (100 kgf peak) at 0.5~m/s (245~W nominal, 490~W peak). This corresponds to dynamically unloading 37~\% of the weight of the heaviest user the multifunctional lift can transfer with a sufficient speed to follow the center of mass of a patient during sit-to-stands \cite{ROEBROECK1994235}, assuming the harness is attached at the center of mass. Current patient assistance devices can unload a weight up to 100~kgf with enough speed to follow the patient in their daily activities \cite{Safegait360} \cite{ErgoTrainer} \cite{ZeroG}. The accuracy of force tracking is also important in this mode. A 5\% tracking error is targeted for the patient not to feel any force variation \cite{HEROUX20051362}.

\subsection{Fall prevention}
Fall prevention is for ensuring safe and comfortable assistance when patients fall during a training session. Since the system is fast, backdrivable, and unable to lift heavy patients in patient assistance mode, injuries may occur if the patient falls. The system must then detect a falling condition and transition to the patient transfer mode in high-speed motion. The maximum acceptable transition height is set to 0.4~m based on the height between the knees and the ground. Furthermore, a 1~m/s² maximum acceleration is set for a comfortable falling deceleration, as it is shown that discomfort starts at an acceleration of 1~m/s² and up to 2~m/s² can be tolerated \cite{Svensson}.

\section{Actuator Design and Modeling} \label{sec:actuator}
This section presents in detail the proposed multifunctional lift. First, different generic design choices are compared to show how the chosen design is superior to more typical design strategies for reaching the same functions. Then, the final actuator design is described, including the experimental prototype and the actual specifications reached. Finally, the equations of motion are detailed.

\subsection{Preliminary Design Exploration}
\begin{figure}[h]
    \centering
    \includegraphics[width = \linewidth]{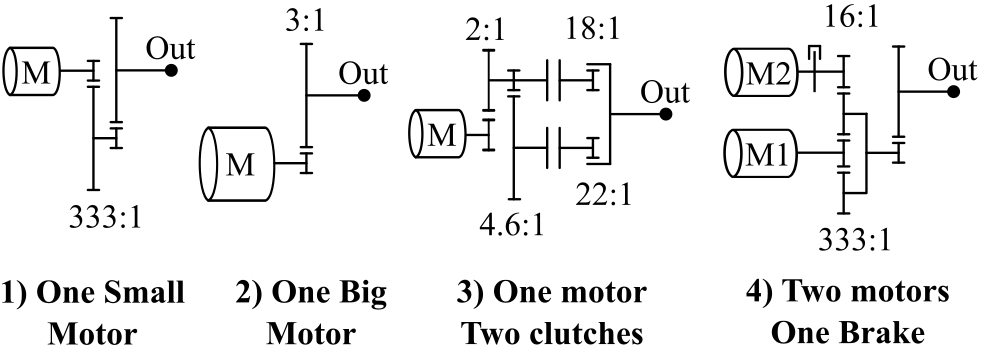}
     \caption{Generic concepts for design comparison, more specifically the ratios of each gear stages based on the requirements and the selected components.}
    \label{fig:schemas_comparaison}
\end{figure}

Figure~\ref{fig:schemas_comparaison} shows alternative concepts possible for a ceiling device based on already explored actuator solutions: 1) one small highly-geared motor \cite{guldmann}; 2) one powerful lightly-geared motor \cite{Safegait360}; 3) one motor, two gear ratios, and two clutches \cite{DCT}; and 4) two motors, two gear ratios, one brake (as proposed).
For each concept, Table~\ref{table:tech_choice2} gives a design proposition based on a list of commercial components and the identification of proper reduction ratios that allows reaching the force requirements for HF and HS modes. Motors are frameless models, so a factor of 1.5 is used for mass, for considering housing mass. It is assumed that by using electromagnetic disk clutches and brakes (which can slip), seamless transitions between HF and HS modes would be possible by controlling the dissipation forces. The mass is for comparison only since it excludes gearboxes, main frame, batteries, etc. Transmission efficiency is assumed 100\% for simplification.

The single small motor design is the lightest (and probably cheapest) design. It has limited dynamics capabilities in HS mode since the inertia and speed would hinder patient accelerations and natural movements. Still, this simple solution may be sufficient for early rehabilitation and aged patients. The single big motor design can reach all force and speed requirements by the use of a powerful motor. The reflected inertia is by comparison low which means better force control for dynamic motions. Mass (and indirectly cost) is however much higher. The selection of the motor size and ratio is a compromise between reflected inertia and motor mass. This system would potentially cause safety hazards due to its high power capabilities. Finally, the last two dual-speed designs can reach all requirements without the need for powerful motors. The fourth design has a better mass/inertia advantage because: 1) disk brakes are lighter than disk clutches (36\% lighter for the MikiPulley models) and 2) for the third design, the reflected inertia of the HF clutch strongly affects the total inertia at the output for the HS mode ($R_1$ being large), which means there is a mass vs. inertia compromise when choosing the reduction ratio between the HF clutch and the motor. One drawback of the two-speed designs is that they require more components, especially more gear stages. The fourth design offered the best tradeoff between performance and size, and was thus chosen for the prototype.

\begin{table}[h]
\vspace{5pt}
\begin{threeparttable}
    \centering
    \caption{Actuator technology comparison}
    \label{table:tech_choice2}
    \begin{tabular}{l c c c c} 
         \hline
         & Small  & Big  & Dual & Dual \\
         & motor$^1$ & motor$^2$ & clutches$^3$ & motors$^4$ \\
         \hline\hline
         HF Force (kgf) & 272 & 272 & 272 & 272 \\
         HS Speed (m/s)& \textbf{0.14} & 0.59 & 0.68 & 0.81 \\
         Reflected Inertia (kg) & \textbf{389} & 14 & 16 & 8 \\
         Mass (kg) & 0.1 & \textbf{5.0} & 1.0 & 0.8 \\
         Transition & N/A & N/A & Yes & Yes \\ 
         \hline
    \end{tabular}
     \begin{tablenotes}
      \footnotesize
      \item $^1$$R=333$, motor Robodrive ILM-E50x08
      \item $^2$$R=3$, motor Tecnotion QTR-A-160-60
      \item $^3$Motor Robodrive ILM-E50x14 $|$ HF line: $R_1=201$, clutch MikiPulley 101-06-13G $|$ HS line: $R_2=37$, clutch MikiPulley 102-05-15
      \item $^4$HF line: $R_1=333$, motor Robodrive ILM-E50x08 $|$ HS line: $R_2=16$, motor Robodrive ILM-E70x18, brake MikiPulley 112-05-12.
    \end{tablenotes}
    \end{threeparttable}
\end{table}



\subsection{Final Design Choice and Prototype}
\begin{figure}[h]
    \centering
    \includegraphics[width=\linewidth]{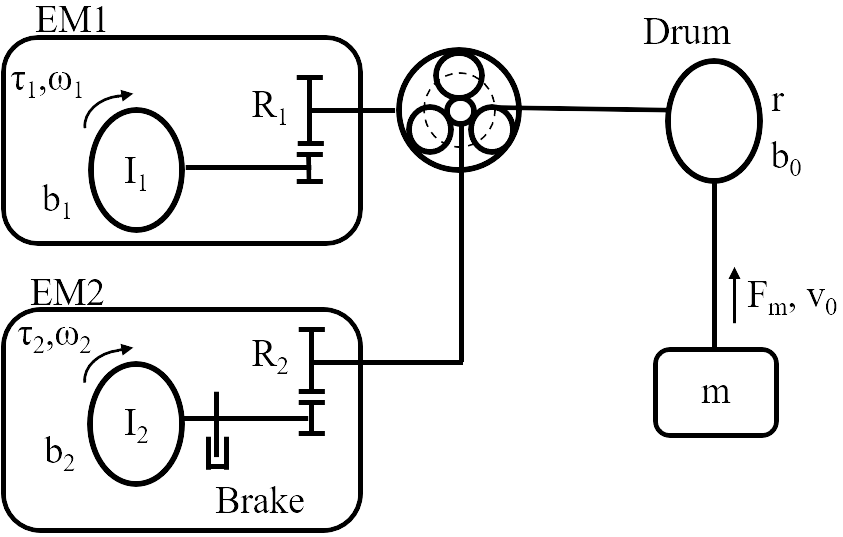}
     \caption{Simplified model of the multifunctional lift using a lumped-parameter approach.}
    \label{fig:schemas}
\end{figure}

The two-speed architecture chosen consists of two motors and a planetary gear train. One motor is a highly geared electric motor (EM1), while the other is a lightly geared electric motor (EM2). As shown in Figure~\ref{fig:schemas}, the drum output that is attached to the patient connects to the carrier of the planetary gearbox. 
The planetary is thus used as a differential. This results in a redundant system, the displacement of each motor adds up to drive the output. 

This architecture has two operating modes, a high-force (HF) mode and a high-speed (HS) mode. 
In HF mode, the brake is closed and EM1 drives the output, resulting in slow displacement but providing high-force capabilities. In HS mode, the brake is opened and EM2 drives the output, resulting in high-speed capabilities and low reflected inertia, which is beneficial for patient assistance during daily activities that require fast movements.
Controlling brake slippage ensures seamless transitions between both control modes of the system by means of high power dissipation capabilities, i.e., high forces at high speed. This provides even more flexibility in meeting the requirements for patient transfer and assistance. For instance, when a patient is falling at high speed, the system needs to apply a large force to stop the patient.

\begin{figure}
    \centering
    \vspace{5pt}
    \includegraphics[width=\linewidth]{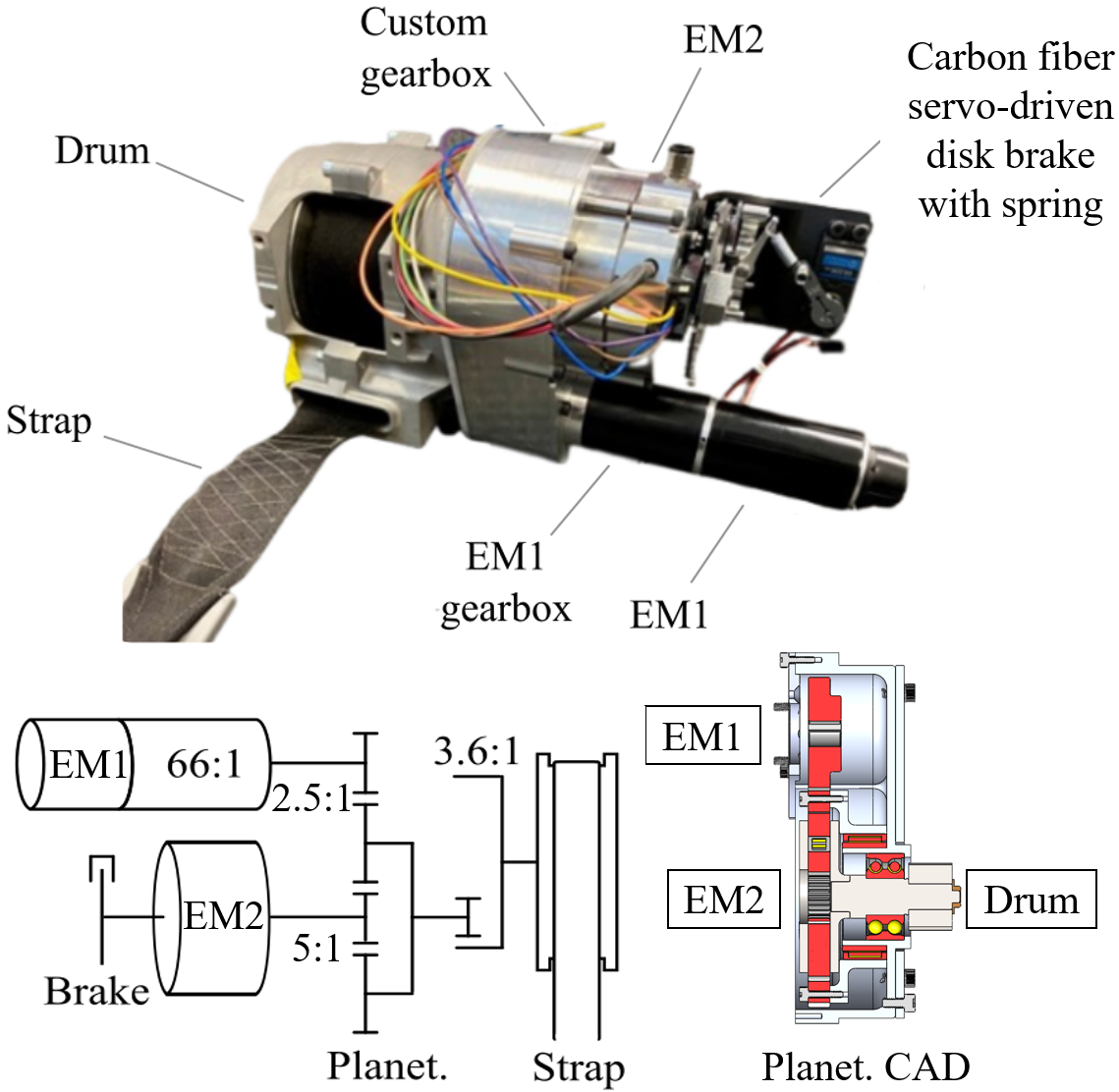}
    \caption{Multifunctional lift experimental prototype and scheme of the actual gearing topology.}
    \label{fig:lM2s}
\end{figure}

A prototype of a multifunctional lift was assembled, including the dual-motor actuator and a drum that winds a strap as pictured in Figure~\ref{fig:lM2s}. The prototype includes a support for a harness that will be installed on a patient. For EM1, a Maxon RE40 150~W brushed motor
with a 66:1 integrated gearbox 
is coupled to the ring gear of the planetary gearbox. For EM2, a Tecnotion QTR-A-78-25 
is coupled directly to the sun gear. 
The custom gearbox includes a planetary gear stage and other spur gear stages. The total reduction ratios are respectively $R_1=600$ and $R_2=18$ for EM1 and EM2. The output of the custom gearbox is connected to a drum (average 0.04~m radius $r$). Encoders are implemented on both motors, and a load cell measures the load applied on the strap. 
EM1 is powered by an Escon50/5 driver and EM2 with an ODriveV3.6 driver. Both motor drives allow for current and velocity control at low level. 
Since EM1 is much stronger than EM2, a disk brake is connected to EM2 to stop it from being backdriven by EM1 during HF mode and for fall prevention. When the disk is in braking position, it locks the sun gear of the gearbox. The disk brake is made of carbon fiber to minimize the reflected inertia in HS mode. The spring is in series between the servomotor (DS3235SG, 35 kg) and the brake caliper (series-elastic actuation) so that the position of the servomotor roughly controls the braking torque. An Arduino Mega is used for the software control with a 250~Hz control loop. 

The prototype's theoretical capabilities are outlined in Table \ref{table:specs}. In HF mode, the load is limited by the 66:1 gearbox rated torque. In HS mode, the peak force is reached when sending twice the nominal current of the motor. All specifications for speed and force in HF and HS modes were reached. The calculated reflected inertia at output includes the motor frameless rotor (59\%), the custom shaft (22\%), and the disk brake (19\%). The lightly-geared design is thus expected to track sit-to-stands and normal walking properly. It also leads to a reasonable reflected inertia (as opposed to the highly-geared line) relative to the nominal assistive forces and patient's weight, and the measured backdriving force is 3.2~kg.



\begin{table}[h]
    \centering
    \caption{Actuator capabilities at the output compared with other similar designs}
    \begin{tabular}{l c c c c c c}
        \hline
        & \multicolumn{2}{c}{Prototype} & \multicolumn{2}{c}{Guldmann} & \multicolumn{2}{c}{Aretech} \\ \hline
        \hline
        Mode & HF & HS & HF & HS & HF & HS \\ \hline
        Ratio & 600:1 & 18:1 & N/A & N/A & N/A & N/A  \\ \hline
        Max force & 318 & 59 & 400 & 100 & 136 & 68\\
        (kgf) &  & (100 peak) & & & &\\ \hline
        Loaded & 0.05 & 0.55 & 0.06 & 0.1 & N/A & N/A \\ 
        velocity (m/s) & & (0.34@peak) & & & &\\ \hline
        Reflected & 3427 & 5.1 & N/A & N/A & N/A & N/A\\ 
        inertia (kg) & & & & & &\\ \hline
    \end{tabular}
    \label{table:specs}
\end{table}

The relationship between the servo's angle and the brake's force was experimentally determined by setting the servo to specific angles and measuring the corresponding output force using a load cell. 
A linear fit was used for later controlling purposes.

\subsection{Equations of Motion}

\begin{table}[h]
    \centering
    \caption{Design parameters summary for both motors and the output}
    \begin{tabular}{l c c c c c}
        \hline
         & Effort & Velocity & Damping & Inertia & Ratio  \\ 
        \hline\hline
        EM1 & $\tau_1$ & $\omega_1$ & $b_1$ & $I_1$ & $R_1$ \\ \hline
        EM2 & $\tau_2$ & $\omega_2$ & $b_2$ & $I_2$ & $R_2$ \\ \hline
        Output & $F_m$ & $v_0$ & $b_0$ & $m$ & $r$ \\ \hline
    \end{tabular}
\end{table}

Since both motors work to drive the output, the kinematic equation of the system can be simplified using a lumped-parameter approach, which leads to:
\begin{equation}\label{equ:kinematic}
    v_0/r = \omega_1/R_1 + \omega_2/R_2 
\end{equation}

where $v_0$ is the linear velocity of the output strap, and $\omega_1$ and $\omega_2$ are the angular velocity of EM1 and EM2. The radius of the drum $r$ is assumed to be constant, and $R_1$ and $R_2$ are the total gear ratios between each motor and the output of the drum, including any gear ratio in the drum, the planetary gearbox, and the motor gearbox. 
The static force relationship is:
\begin{equation}
    F_m r = R_1 \tau_1 = R_2 (\tau_2 - \tau_B sign(\omega_2)) 
\end{equation}

where $F_m$ represents the output force on the drum, $\tau_1$ and $\tau_2$ the motor torques of EM1 and EM2. $\tau_B$ is the friction torque of the brake. When considering the inertial properties of the system, a 2 degrees of freedom state space model can be constructed with the states being $v_0$ and $\omega_1$:
\begin{equation}
     \begin{bmatrix} \dot{v_0} \\ \dot{w_1} \end{bmatrix} \hspace{-2pt} = \hspace{-2pt} \begin{bmatrix} -H^{-1} D \end{bmatrix} \hspace{-2pt} \begin{bmatrix} v_0 \\ w_1 \end{bmatrix} \hspace{-2pt} + \hspace{-2pt} \begin{bmatrix} H^{-1} B \end{bmatrix} \hspace{-2pt} \begin{bmatrix} \tau_1 \\ \tau_2 - \tau_B sign(\omega_2)  \\ F_m \end{bmatrix}
\end{equation}

with:
\begin{equation}
    H = \begin{bmatrix} m + \cfrac{R_2^2 I_2}{r^2} & -\cfrac{R_2^2 I_2}{R_1 r}  \\ -\cfrac{R_2^2 I_2}{R_1 r} & I_1 + \cfrac{R_2^2 I_2}{R_1^2} \end{bmatrix}   
\end{equation}
\begin{equation}
    D = \begin{bmatrix} b_0 + \cfrac{R_2^2 b_2}{r^2} & -\cfrac{R_2^2 b_2}{R_1 r}  \\ -\cfrac{R_2^2 I_2}{R_1 r} & b_1 + \cfrac{R_2^2 b_2}{R_1^2} \end{bmatrix}
    B = \begin{bmatrix} 0 & \cfrac{R_2}{r} & -1  \\ 1 & -\cfrac{R_2}{R_1} & 0 \end{bmatrix}
\end{equation}

with $m$ being the load on the strap and including the inertia of the drum, and $I_1$ and $I_2$ the inertia of motors EM1 and EM2. $b_0$, $b_1$ and $b_2$ represent the linear viscous friction terms of the output, EM1 and EM2.

In HF mode, the brake is fully closed and only EM1 can drive the output. Hence, the equation can be simplified to a single motor, single output system: 
\begin{equation}
     \left (m + I_1 \cfrac{R_1^2}{r^2} \right) \dot{v_0} = \cfrac{R_1}{r} (\tau_1 - b_1 \omega_1) - b_0 v_0 - F_m
\end{equation}

For HS mode, both motors can contribute to the motion of the output. However, since EM1 has a greater mechanical advantage over EM2, the equation can be simplified when $R_1 >> R_2$, which is the case for the prototype by design: 
\begin{equation}
    \left(m + I_2 \cfrac{R_2^2}{r^2} \right) \dot{v_0} =
    \cfrac{R_2}{r}(\tau_2 - \tau_B sign(\omega_2) - b_2 w_2)  
    - b_0 v_0      
    - F_m
\end{equation}

The influence of EM1 in HS mode is thus negligible on the motion of the output, but still affects the motion of EM2 from the kinematic equation \ref{equ:d_kinematic} when working with the nullspace and exploiting the redundant degree of freedom of the system \cite{Girard2015}:

\begin{equation}\label{equ:d_kinematic}
    \dot{\omega_2} = R_2 ( \dot{v_0}/r -\dot{\omega_1}/R_1 )
\end{equation}

All in all, the equations for both modes have the same structure and the differences come from the relation between the mechanical advantage $R_1$ over $R_2$.

\section{CONTROL ALGORITHMS} \label{sec:algo}

\begin{figure}[h]
    \centering
    \includegraphics[width=\linewidth]{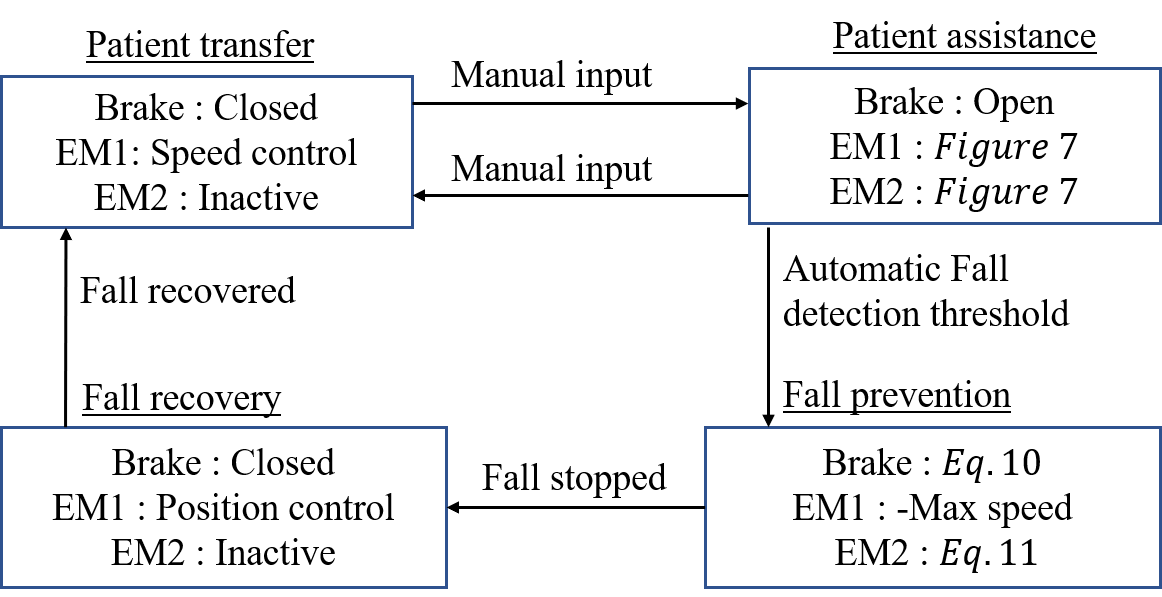}
    \caption{Mode state machine with transfer mode for high force capabilities, assistance mode for high-speed capabilities, fall prevention to brake the patient, and fall recovery to lift the patient.}
    \label{fig:stateMachine}
\end{figure}

The system's main operating modes are high-force mode and high-speed mode, which can be changed manually as shown in Figure \ref{fig:stateMachine}. The multifunctional lift also needs to change mode automatically from HS to HF if the patient were to fall, to support the full weight of the patient and stop the fall. 

HF and HS mode transitions occur by opening or closing the brake. If a fall is detected in HS mode, the brake and EM2 work simultaneously to decelerate the patient before downshifting. 
To help downshift faster, EM1 can accelerate downward to slow down EM2 (see equation \ref{equ:d_kinematic}).
Once the velocity of EM2 is zero, the brake closes to its maximal braking force and the system becomes in HF mode. To help the patient recover and continue his activity, the system then lifts them to the position where the fall was detected. In the following section, controller details are presented for each operating mode. 

\subsection{Patient Transfer}
To transfer a patient, the multifunctional lift needs to hoist the full weight of the patient. This requires the system to be in HF mode utilizing the highly geared motor. The controller is a velocity command sent to EM1 based on a user up-down input on a remote. This is the base functionality of currently available transfer lifts.

\subsection{Patient Assistance}

\begin{figure}[h]
    \centering
    \includegraphics[width=\linewidth]{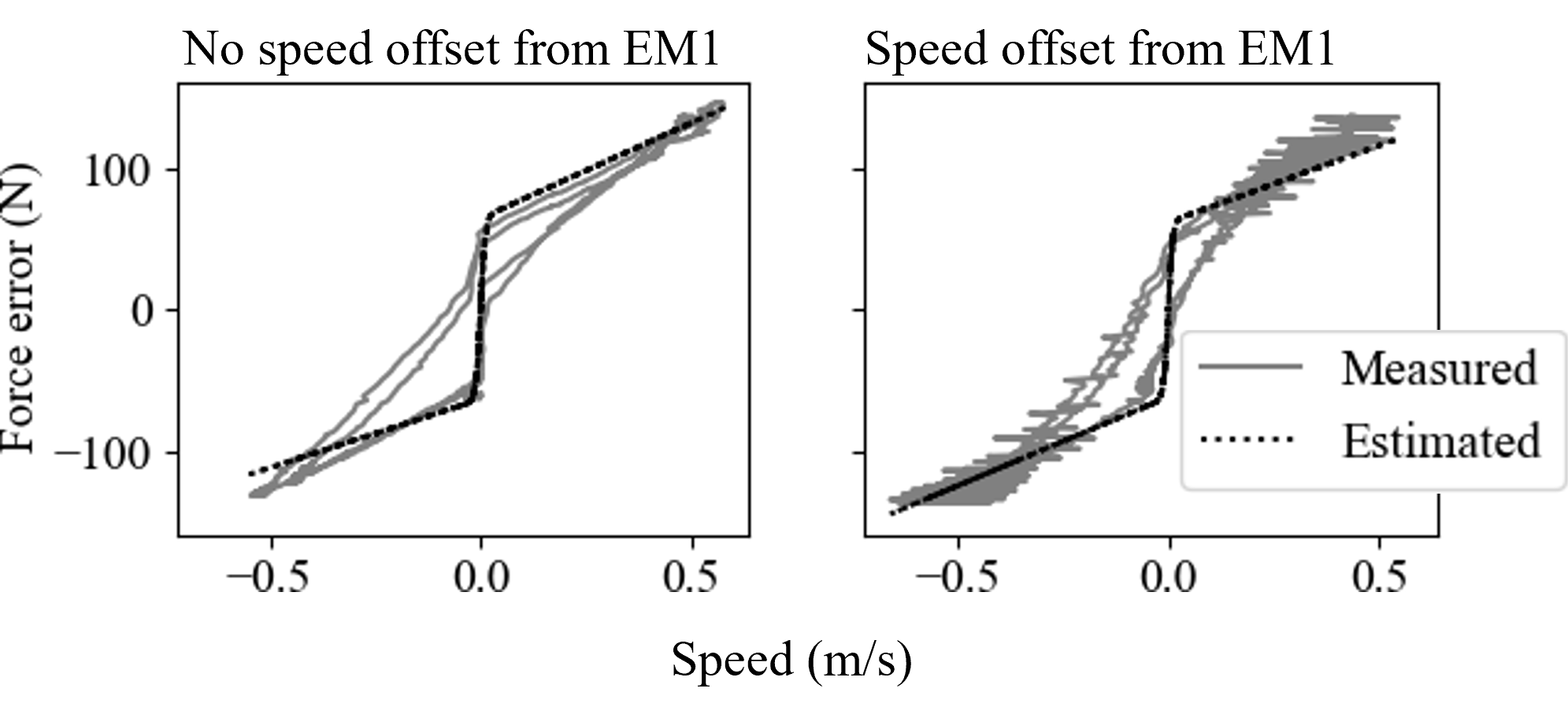}
    \caption{Measure of the friction in EM2 without and with speed offset from EM1 while a 150~N load is applied.}
    \label{fig:friction}
\end{figure}

For daily activities assistance, the lift unloads weight to the patient while following their movements. This requires the multifunctional lift to be in HS mode and to control the force at the output rather than the position or velocity as in transfer mode. 
There are different ways to control the output force of an actuator. 
a) The output torque can be obtained by knowing the motor constant ($k_t$) and adjusting the current of EM2. This works well for quasi direct drive motors \cite{QDD_bipedal}, it results in a simpler controller and removes the need for a load cell. 
However, since EM2 is a lightly geared motor, the friction is augmented by the gearbox and can induce an undesired error on the force output. 
This error can be measured in terms of speed (see Figure \ref{fig:friction}) and a dry and viscous friction model \cite{Bona} can be estimated using a linear equation (using the absolute value for symmetry) multiplied by a hyperbolic tangent function:
\begin{equation}
    \tau_f = f(\omega_2) = ( b |\omega_2| + c + d F_d) tanh(\omega_2)
\end{equation} 

Where b is the viscous friction coefficient, c is the dry friction offset and d a scaling parameter to offset the dry friction with the desired force.

b) The friction estimation can then be subtracted to the desired force to compensate for the friction error and help reduce the force tracking error. One issue with friction compensation is when trying to compensate near zero speed since a more complex friction model would be necessary for considering hysteresis. c) With a dual-motor actuator, EM1 can run at a constant speed while EM2 is applying a constant force. This offsets the speed of EM2 from the output speed, which reduces the time EM2 is at zero speed (see equation~\ref{equ:kinematic}), since the patient is often near zero speed (when they're standing still or sitting). 
Research prototypes for assistance devices often use the feedback from a load cell to measure the force applied on the user and send it to a closed loop controller, for instance d) a PID \cite{aretech} or e) a disturbance observer \cite{DOB_BWS}. 

\begin{figure}[h]
    \centering
    \includegraphics[width=\linewidth]{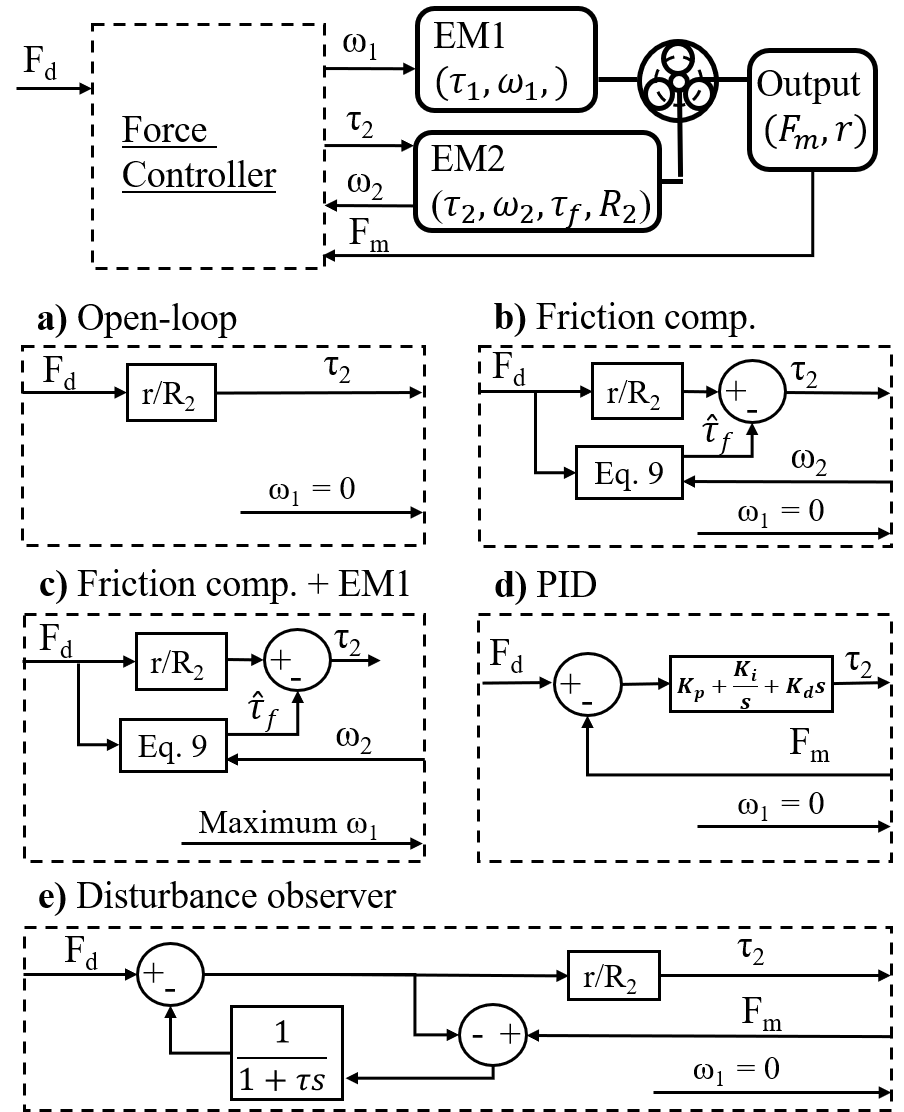}
    \caption{Control architecture and force controllers. Where a) is an open-loop current control, b) a friction compensation algorithm, c) a friction compensation algorithm with an offset on the speed, d) a PID current controller, and e) a DOB}
    \label{fig:controller}
\end{figure}

The force control algorithms that were tested are presented in Figure~\ref{fig:controller}: a) an open-loop current control, b) a friction compensation algorithm, c) a friction compensation algorithm with an offset on the speed by leveraging the nullspace, d) a PID current controller and e) a disturbance observer (DOB).

\subsection{Fall prevention and recovery}


There are two goals to the fall prevention algorithm: first, stopping the fall of a patient safely and comfortably, and second, helping the patient resume their daily activity afterward.
To ensure the safety of the patient, the multifunctional lift must never let the patient reach the ground during a fall. As for the comfort, it is dependent on the patient's acceleration. To apply a desired deceleration ($a_d$) to the patient, the torque needed from the brake can be calculated knowing the weight of the patient (see equation \ref{equ:brake}), which is measured from the load cell present in the system.

\begin{equation}
    \tau_B = \cfrac{r}{R_2}m(g + a_d)
    \label{equ:brake}
\end{equation}

\begin{equation}
    \tau_2 = k(\int a_d dt - v_0)
    \label{equ:fallEM2Eq}
\end{equation}

Due to servo control, the brake applies a discrete amount of force with some delay as it reaches the desired angle. To improve the accuracy and response time of the system, EM2 applies a torque proportional to the error on the desired speed as shown by equation \ref{equ:fallEM2Eq}, where $k$ is a proportional gain.

\section{EXPERIMENTS} \label{sec:exp}

\begin{figure}[h]
    \centering
    \includegraphics[width=\linewidth]{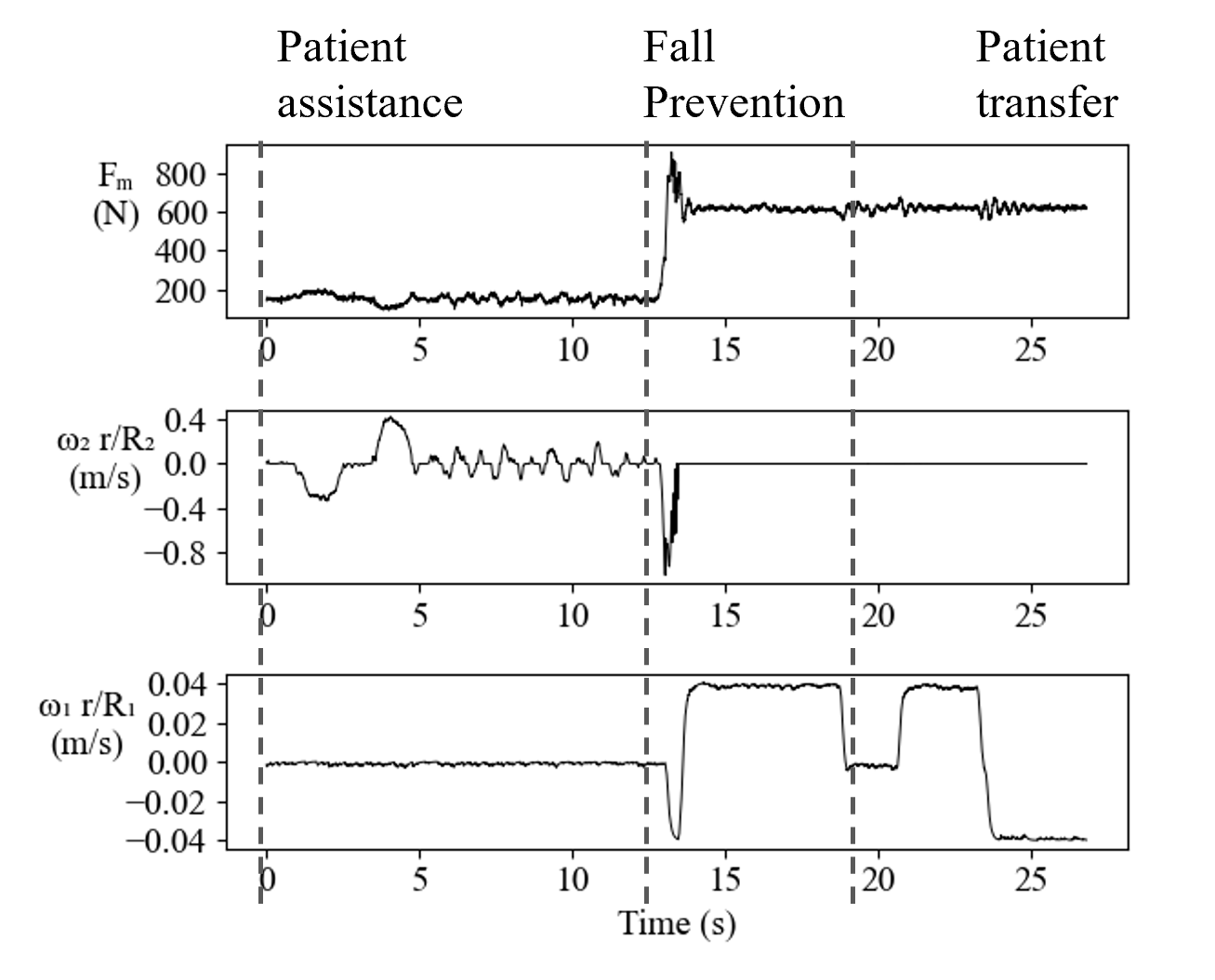}
    \caption{Complete trial that encompasses patient assistance, fall prevention, and patient transfer, showing the force measured, and the speed contribution of EM2 and EM1.}
    \label{fig:fulltrial}
\end{figure}

To better comprehend the functionality of the system, Figure~\ref{fig:fulltrial} shows a complete trial that includes all three modes of operation (patient assistance, fall prevention, and patient transfer). Trials for patient assistance and fall prevention are described in more details in the following paragraphs.

During patient assistance, a 150~N unloading is applied along with the disturbance observer algorithm. The patient starts from a standing position, sits down, stands back up, walks a few steps, and then simulates a fall. The multi-functional lift detects the fall and, the brake and EM1 are activated to stop the patient. Once the speed of EM2 reaches 0~m/s, EM1 can lift the patient to the height at which the fall was detected. The patient is now fully supported by the multi-functional lift, and EM1 can move the patient in transfer mode. 
The following section presents experimental validations for the patient assistance and fall prevention algorithms.

\subsection{Force fidelity}

\begin{figure}[h]
    \centering
    \includegraphics[width=\linewidth]{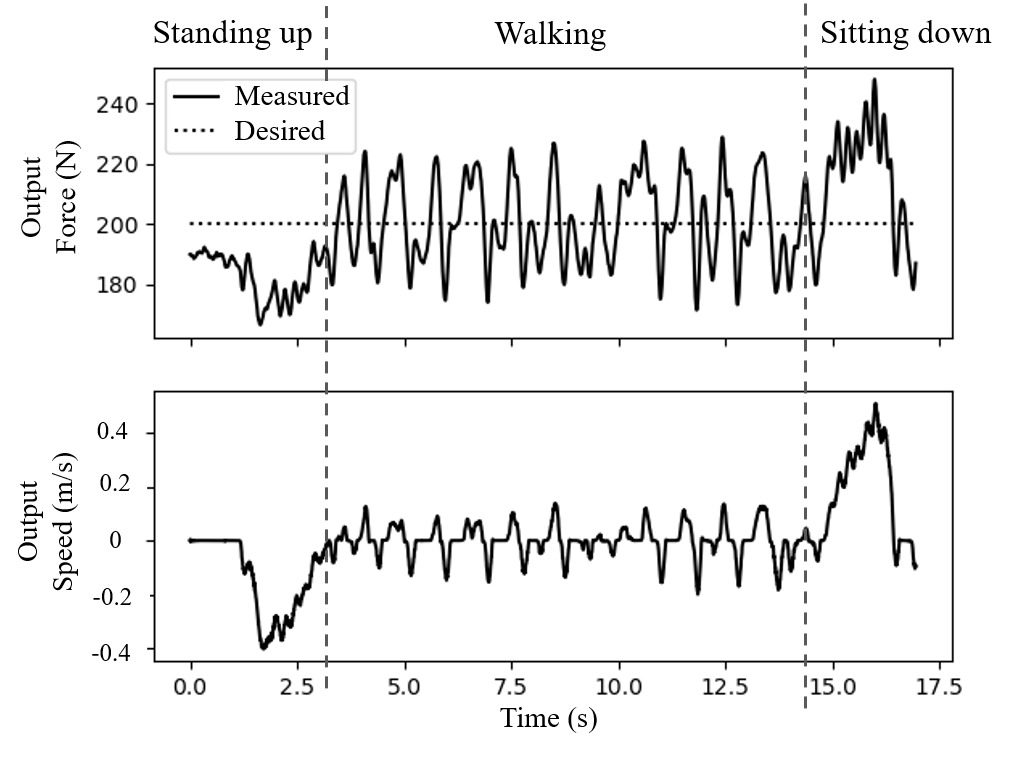}
    \caption{Force and speed during a fixed course with friction compensation algorithm b).}
    \label{fig:course}
\end{figure}

Experiments were done to compare the different patient assistance control algorithms. The quality of force is measured with eight healthy adults\footnote{All tests were approved at Université de Sherbrooke by the \textit{ Comité d’éthique de la recherche – Lettres et sciences humaines} on Dec. 13 2022} (age: from 22 to 29 years old, weight: from 45 to 100 kg, 2 females and 6 males) 
on a predetermined course (see Figure \ref{fig:course}). The user starts from a sitting position, stands up, walks for four meters, and sits back down, similarly to the “get-up and go” \cite{Laustsen-Kiel2021-io} test, while the system unloads 200~N which was determined as a comfortable unloading of force to the users. The experiment is repeated with the different control algorithms. 
The force applied to the output is measured using the strap's load cell during experiments. Mean absolute error is calculated and compared for standing up, sitting down, and walking phases. The speed is also measured using the encoders on the motors to separate the walking and the standing/sitting phases. The system moves at higher speeds, above 0.3~m/s during the standing and sitting phases. For the walking phase, the system's speed is lower than 0.3~m/s (see Table \ref{table:Force_algo_results}). The friction compensation with EM1 algorithm is then compared to every other algorithm using a Mann-Whitney test.

\begin{table}[h!]
    \centering
    \caption{Absolute mean error of force (N) for the different algorithms of Figure \ref{fig:controller}}
    \begin{tabular}{c c c c c c} 
         \hline
         Phase & \multicolumn{4}{c}{Absolute mean force tracking error (N)}\\
         \hline
         & Current & Friction  & Friction & PID  & DOB \\
         & a) & comp. b) & comp. with & d)& e)\\
         & & & EM1 c)& & \\
         \hline\hline
         Standing/Sitting & 104.5 & 38.4 & 24.9 & 29.7 & 34.7 \\         
         Walking & 45.5 & 34.8 & 15.7 & 16.1 & 15.3 \\ 
         \hline
    \end{tabular}
    \label{table:Force_algo_results}
\end{table}

As mentioned, controlling force with a constant current leads to errors due to system friction. Although, compensating for the friction improves the force tracking accuracy, at lower speeds the friction is difficult to compensate and the error is similar to that observed with a constant current. However, the use of EM1 to offset the zero speed improves the force tracking accuracy even further when the system is at low speed. The proposed algorithm is then compared with a traditional PID controller and a disturbance observer, utilizing force feedback from a load cell. At high speeds, the friction compensation gives similar performances to other feedback algorithms, but at lower speeds, the performances are similar to a constant current. At low speed, the average tracking error for the friction compensation algorithm with speed offset is 15.7~N, which corresponds to 7.8\% of the nominal assistance force while at high speed it is 24.9~N or 12\%, both of which are higher than the 5\% target for the assistance force, but the general feeling the participants felt, which was not measured, was said good.

\subsection{Fall prevention and recovery}

\begin{figure}[h]
    \centering
    \includegraphics[width=\linewidth]{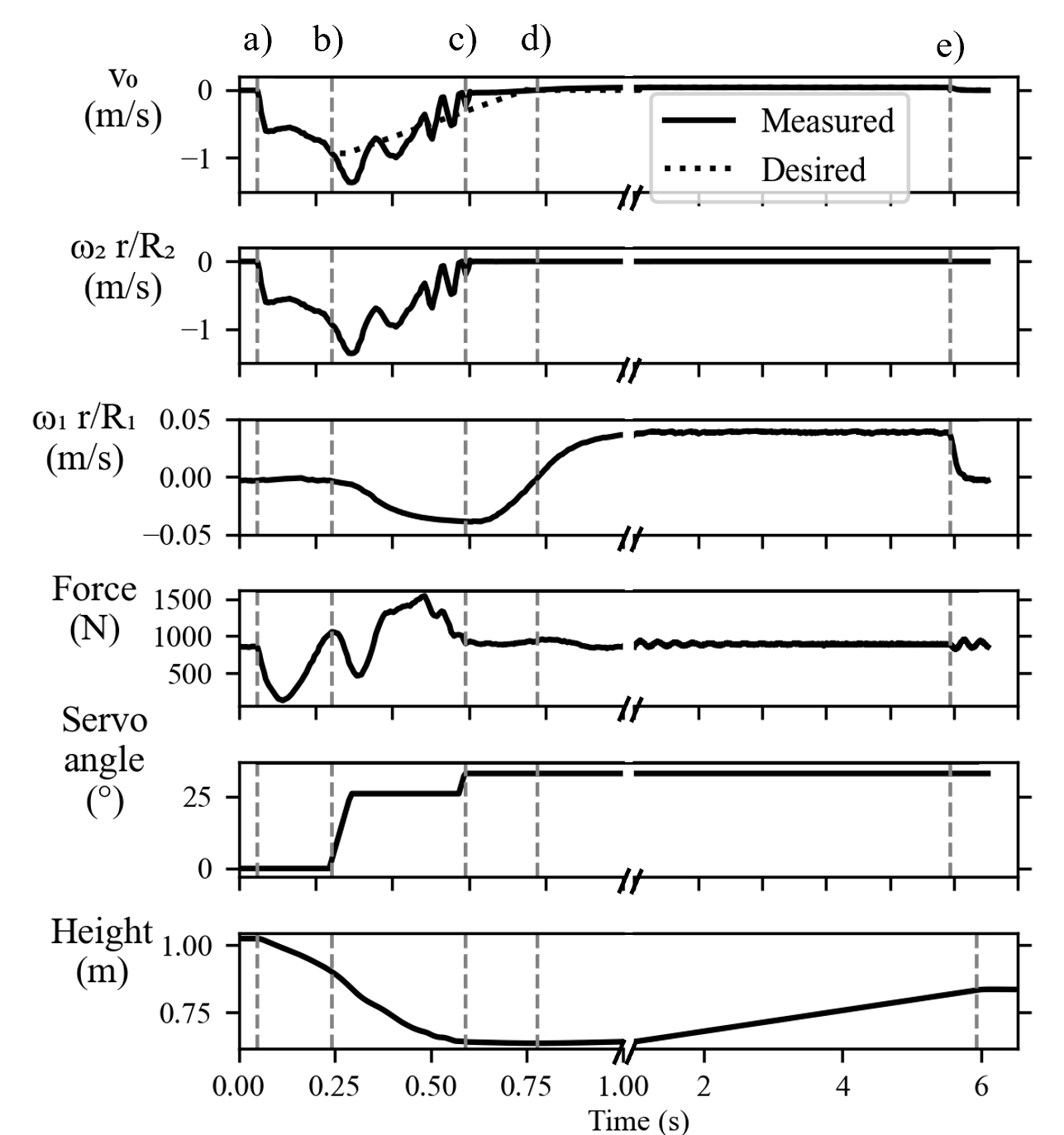}
    \caption{Complete fall sequence with the output speed, the speed contribution of EM2 and of EM1 (see equation. \ref{equ:kinematic}), the output force, the servo angle, and the output position. The different events are: a) start of the fall, b) start of the braking sequence, c) EM2 stops, d) EM1 stops, e) end of patient recovery.}
    \label{fig:fallSequence}
\end{figure}

To validate the performances of the brake during fall prevention, tests with different weights (68~kg, 90~kg, 113~kg) were conducted with a desired deceleration of $1~m/s^2$, $2~m/s^2$ and the maximum possible braking force to find the minimal fall distance.
Figure \ref{fig:fallSequence} shows the fall sequence of a test with a 90~kg weight at $2~m/s^2$. a) Starting at a height of 1~m, the weight gradually falls until, b) the encoder in EM2 reads an output speed of 0.90~m/s, triggering the system to engage the brake for controlled deceleration. To help downshift faster, EM1 applies its maximum downward velocity (see equation \ref{equ:kinematic}). c) Once EM2 stops, EM1 takes over to slow the fall until, d) it stops completely. e) Finally, since only EM1 can lift the full weight of the patient, EM1 lifts the patient to the height at which the fall was detected to facilitate the patient's ability to stand up and resume their activity.

By measuring the height at the detection of the fall and at the lowest point, it is possible to find the fall distance $x$. The average acceleration during the fall is calculated with $a = v_i^2/(2x)$ with $v_i$ being the velocity of the mass at the detection of the fall, and the average force is found with $F = m(g + a)$. Table \ref{table:fall} shows the results of the tests conducted. Measures for the average acceleration, fall distance and average forces were compared to their theoretical values. The acceleration was the parameter that was adjusted and is related to the patient's comfort. The fall distance is more intuitive to understand and is linked to the patient's safety since it indicates the minimal falling distance that the patient needs. The force roughly corresponds to the braking capabilities of the system.

\begin{table}[h!]
    \centering
    \caption{Fall comparison tests}
    \begin{adjustbox}{width= \linewidth}
    \begin{tabular}{c c c c c c c c}
        \hline
        & & \multicolumn{3}{c}{Theoretical} & \multicolumn{3}{c}{Measured}\\
        \cline{3-8}
        Weight & Test & Acc. & Dist. & Force & Acc. & Dist. & Force \\
        (kg)& & $(m/s^2)$ & (m) & (N) & $(m/s^2)$ & (m) & (N) \\
        \hline\hline
        & Max. force & 4.2 & 0.10 & 951.0 & 4.3 & 0.09 & 961.1 \\
        68 & 2 $m/s^2$ & 2.0 & 0.20 & 803.1 & 2.3 & 0.18 & 823.0 \\
        & 1 $m/s^2$ & 1.0 & 0.41 & 735.1 & 1.5 & 0.28 & 767.4 \\
        \hline
        & Max. force & 3.7 & 0.12 & 1191 & 2.3 & 0.18 & 1090 \\
        90 & 2 $m/s^2$ & 2.0 & 0.20 & 1062 & 1.6 & 0.26 & 1022 \\
        & 1 $m/s^2$ & 1.0 & 0.41 & 972.2 & 1.4 & 0.30 & 1005 \\
        \hline
        & Max. force & 2.8 & 0.15 & 1424 & 1.6 & 0.25 & 1290 \\
        113 & 2 $m/s^2$ & 2.0 & 0.20 & 1334 & 1.2 & 0.34 & 1243 \\
        & 1 $m/s^2$ & 1.0 & 0.41 & 1221 & 0.9 & 0.45 & 1209 \\
        \hline
    \end{tabular}
    \end{adjustbox}
    \label{table:fall}
\end{table}

At the beginning of the braking sequence, only EM2 applies a torque to slow the mass. Since EM2 can only lift up to 59~kgf, with lighter patients, EM2's contribution to braking torque is significant relative to the patient's weight, allowing the system to apply the desired braking torque from the beginning without brake input. This means the system can decelerate the mass on a shorter distance. With heavier masses, EM2's influence diminishes, requiring more braking distance.
Patients under 90~kg can be safely decelerated (at 1~m/$s^2$) and stopped safely (within 0.40~m). Heavier patients require higher deceleration to ensure safety. With heavier patients, it is expected that the current brake would be unable to stop their fall on less than 0.40~m. Thus, a product redesign would require a faster, stronger brake for patient safety.

\section{CONCLUSIONS}

This paper introduces a prototype ceiling robot capable of patient transfer, assistance, and fall prevention using a dual-motor actuator. The prototype lifts patients up to 318~kg at 0.05~m/s in transfer mode and unloads up to 59~kg at 0.55~m/s in assistance mode. 
During patient assistance, force tracking error was 12\% at high speed and 7.8\% at low speed using a friction compensation algorithm and both motors, comparable to a traditional closed-loop controller. This is over the 5\% error target, but it was only tested with eight healthy adults as a preliminary validation, and no participant mentioned discomfort during the tests.  
Future work involves clinical evaluation with patients of different profiles to validate if the technology meets patient needs. This could determine if the force controller needs further improvement for patient comfort.

For the fall prevention algorithm, the patient's deceleration can be controlled by using the disk brake to apply a desired amount of force to dissipate energy.
To better control the braking of a fall and ensure the safety of all patients, the servo motor actuator could be replaced by a faster and stronger brake. Also, the brake could be normally closed to increase the safety of the patients.
While suitable for patient lifts, the actuators could also serve a wider range of robots with dual operation points and rapid downshifting capabilities.

\bibliographystyle{IEEEtran}
\bibliography{IEEEabrv, ref}

\begin{thebibliography}{10}
\providecommand{\url}[1]{#1}
\csname url@rmstyle\endcsname
\providecommand{\newblock}{\relax}
\providecommand{\bibinfo}[2]{#2}
\providecommand\BIBentrySTDinterwordspacing{\spaceskip=0pt\relax}
\providecommand\BIBentryALTinterwordstretchfactor{4}
\providecommand\BIBentryALTinterwordspacing{\spaceskip=\fontdimen2\font plus
\BIBentryALTinterwordstretchfactor\fontdimen3\font minus
  \fontdimen4\font\relax}
\providecommand\BIBforeignlanguage[2]{{%
\expandafter\ifx\csname l@#1\endcsname\relax
\typeout{** WARNING: IEEEtran.bst: No hyphenation pattern has been}%
\typeout{** loaded for the language `#1'. Using the pattern for}%
\typeout{** the default language instead.}%
\else
\language=\csname l@#1\endcsname
\fi
#2}}

\bibitem{SUMIDA2001391}
M.~Sumida, M.~Fujimoto, A.~Tokuhiro, T.~Tominaga, A.~Magara, and R.~Uchida,
  ``Early rehabilitation effect for traumatic spinal cord injury,''
  \emph{Archives of Physical Medicine and Rehabilitation}, vol.~82, no.~3, pp.
  391--395, 2001.

\bibitem{Dirkes}
\BIBentryALTinterwordspacing
S.~M. Dirkes and C.~Kozlowski, ``{Early Mobility in the Intensive Care Unit:
  Evidence, Barriers, and Future Directions},'' \emph{Critical Care Nurse},
  vol.~39, no.~3, pp. 33--42, 06 2019.
\BIBentrySTDinterwordspacing

\bibitem{Functionalstatus}
O.~Tonkikh, E.~Shadmi, N.~Flaks‐Manov, M.~Hoshen, R.~D. Balicer, and
  A.~Zisberg, ``Functional status before and during acute hospitalization and
  readmission risk identification,'' \emph{Journal of Hospital Medicine},
  vol.~11, no.~9, p. 636–641, Sep 2016.

\bibitem{Motorandcognitive}
A.~Middleton, J.~E. Graham, Y.-L. Lin, J.~S. Goodwin, J.~P. Bettger,
  A.~Deutsch, and K.~J. Ottenbacher, ``Motor and cognitive functional status
  are associated with 30-day unplanned rehospitalization following post-acute
  care in medicare fee-for-service beneficiaries,'' \emph{Journal of General
  Internal Medicine}, vol.~31, no.~12, p. 1427–1434, Dec 2016.

\bibitem{Safegait360}
\BIBentryALTinterwordspacing
Hocoma. Safegait 360° by gorbel. [Online]. Available:
  \url{https://www.hocoma.com/us/solutions/safegait360/product-brochure/}
\BIBentrySTDinterwordspacing

\bibitem{ErgoTrainer}
\BIBentryALTinterwordspacing
Winncare. Ergo trainer. [Online]. Available:
  \url{https://www.winncare.com/fiche-produits-rehabilitation-ergo_trainer-12-630-int.htm}
\BIBentrySTDinterwordspacing

\bibitem{aretech}
J.~Hidler, D.~Brennan, D.~Nichols, K.~Brady, T.~Nef, \emph{et~al.}, ``Zerog:
  overground gait and balance training system.'' \emph{Journal of
  Rehabilitation Research \& Development}, vol.~48, no.~4, 2011.

\bibitem{Handicare}
\BIBentryALTinterwordspacing
Handicare. Fixed ceiling lift motors. [Online]. Available:
  \url{https://www.handicare.ca/product-category/homecare/ceiling-lifts/fixed-ceiling-lifts/}
\BIBentrySTDinterwordspacing

\bibitem{ms2}
\BIBentryALTinterwordspacing
Arjo. Ceiling lifts. [Online]. Available:
  \url{https://www.arjo.com/int/products/patient-handling/ceiling-lifts/}
\BIBentrySTDinterwordspacing

\bibitem{Savaria}
\BIBentryALTinterwordspacing
Savaria. Savaria fl fixed lifts. [Online]. Available:
  \url{https://ceiling-lift.com/product/savaria-fl-fixed-lift/}
\BIBentrySTDinterwordspacing

\bibitem{guldmann}
\BIBentryALTinterwordspacing
Guldmann. Trainer module for gh3+. [Online]. Available:
  \url{https://www.guldmann.com/ca/products/ceiling-lift-systems/lifting-motors/trainer-module-for-gh3plus}
\BIBentrySTDinterwordspacing

\bibitem{ROEBROECK1994235}
\BIBentryALTinterwordspacing
M.~Roebroeck, C.~Doorenbosch, J.~Harlaar, R.~Jacobs, and G.~Lankhorst,
  ``Biomechanics and muscular activity during sit-to-stand transfer,''
  \emph{Clinical Biomechanics}, vol.~9, no.~4, pp. 235--244, 1994.
\BIBentrySTDinterwordspacing

\bibitem{Frey}
M.~Frey, G.~Colombo, M.~Vaglio, R.~Bucher, M.~Jorg, and R.~Riener, ``A novel
  mechatronic body weight support system,'' \emph{IEEE Transactions on Neural
  Systems and Rehabilitation Engineering}, vol.~14, no.~3, pp. 311--321, 2006.

\bibitem{Rad}
M.~H. Rad and S.~Behzadipour, ``Design and implementation of a new body weight
  support (bws) system,'' in \emph{2017 5th RSI International Conference on
  Robotics and Mechatronics (ICRoM)}, 2017, pp. 69--75.

\bibitem{DOB_BWS}
J.~Kwak, W.~Choi, and S.~Oh, ``Modal force and torque control with wire-tension
  control using series elastic actuator for body weight support system,'' in
  \emph{IECON 2017 - 43rd Annual Conference of the IEEE Industrial Electronics
  Society}, 2017, pp. 6739--6744.

\bibitem{GUVENC1994623}
\BIBentryALTinterwordspacing
L.~Güvenç and K.~Srinivasan, ``Friction compensation and evaluation for a
  force control application,'' \emph{Mechanical Systems and Signal Processing},
  vol.~8, no.~6, pp. 623--638, 1994.
\BIBentrySTDinterwordspacing

\bibitem{MacLean}
\BIBentryALTinterwordspacing
M.~K. MacLean and D.~P. Ferris, ``{Design and Validation of a Low-Cost
  Bodyweight Support System for Overground Walking},'' \emph{Journal of Medical
  Devices}, vol.~14, no.~4, p. 045001, 09 2020.
\BIBentrySTDinterwordspacing

\bibitem{lokolift}
M.~Frey, G.~Colombo, M.~Vaglio, R.~Bucher, M.~Jorg, and R.~Riener, ``A novel
  mechatronic body weight support system,'' \emph{IEEE Transactions on Neural
  Systems and Rehabilitation Engineering}, vol.~14, no.~3, pp. 311--321, 2006.

\bibitem{Girard2015}
A.~Girard and H.~H. Asada, ``A two-speed actuator for robotics with fast
  seamless gear shifting,'' in \emph{2015 IEEE/RSJ International Conference on
  Intelligent Robots and Systems (IROS)}, 2015, pp. 4704--4711.

\bibitem{Takayama}
T.~Takayama, T.~Yamana, and T.~Omata, ``Three-fingered eight-dof hand that
  exerts 100-n grasping force with force-magnification drive,'' \emph{IEEE/ASME
  Transactions on Mechatronics}, vol.~17, no.~2, pp. 218--227, 2012.

\bibitem{lecavalier2022bimodal}
\BIBentryALTinterwordspacing
A.~Lecavalier, J.~Denis, J.-S. Plante, and A.~Girard, ``A bimodal hydrostatic
  actuator for robotic legs with compliant fast motion and high lifting
  force,'' \emph{Actuators}, vol.~12, no.~12, 2023.
\BIBentrySTDinterwordspacing

\bibitem{Denis2022}
J.~Denis, A.~Lecavalier, J.-S. Plante, and A.~Girard, ``Multimodal hydrostatic
  actuators for wearable robots: A preliminary assessment of mass-saving and
  energy-efficiency opportunities,'' in \emph{2022 International Conference on
  Robotics and Automation (ICRA)}, 2022, pp. 8112--8118.

\bibitem{VERSTRATEN2018134}
\BIBentryALTinterwordspacing
T.~Verstraten, R.~Furnémont, P.~López-García, D.~Rodriguez-Cianca, H.-L.
  Cao, B.~Vanderborght, and D.~Lefeber, ``Modeling and design of an
  energy-efficient dual-motor actuation unit with a planetary differential and
  holding brakes,'' \emph{Mechatronics}, vol.~49, pp. 134--148, 2018.
\BIBentrySTDinterwordspacing

\bibitem{ZeroG}
\BIBentryALTinterwordspacing
AretechLLC. Zero g gait \& balance system. [Online]. Available:
  \url{https://www.aretechllc.com/products/zerog-gait-and-balance/#product-tech-specs}
\BIBentrySTDinterwordspacing

\bibitem{HEROUX20051362}
\BIBentryALTinterwordspacing
M.~E. Héroux and F.~Tremblay, ``Weight discrimination after anterior cruciate
  ligament injury: A pilot study,'' \emph{Archives of Physical Medicine and
  Rehabilitation}, vol.~86, no.~7, pp. 1362--1368, 2005.
\BIBentrySTDinterwordspacing

\bibitem{Svensson}
\BIBentryALTinterwordspacing
L.~Svensson and J.~Eriksson. Tuning for ride quality in autonomous vehicle~:
  Application to linear quadratic path planning algorithm (dissertation).
  [Online]. Available:
  \url{http://urn.kb.se/resolve?urn=urn:nbn:se:uu:diva-257387}
\BIBentrySTDinterwordspacing

\bibitem{DCT}
\BIBentryALTinterwordspacing
P.~Walker, B.~Zhu, and N.~Zhang, ``Powertrain dynamics and control of a two
  speed dual clutch transmission for electric vehicles,'' \emph{Mechanical
  Systems and Signal Processing}, vol.~85, pp. 1--15, 2017.
\BIBentrySTDinterwordspacing

\bibitem{QDD_bipedal}
Y.~Zhao, S.~Lin, Z.~Zhu, and Z.~Jia, ``A bipedal wheel-legged robot with
  high-frequency force control by qausi-direct drive: Design and experiments,''
  in \emph{2022 IEEE International Conference on Robotics and Biomimetics
  (ROBIO)}, 2022, pp. 58--63.

\bibitem{Bona}
B.~Bona and M.~Indri, ``Friction compensation in robotics: an overview,'' in
  \emph{Proceedings of the 44th IEEE Conference on Decision and Control}, 2005,
  pp. 4360--4367.

\bibitem{Laustsen-Kiel2021-io}
C.~M. Laustsen-Kiel, E.~Lauritzen, L.~Langhans, and T.~Engberg~Damsgaard,
  ``\BIBforeignlanguage{en}{Study protocol for a 10-year prospective
  observational study, examining lymphoedema and patient-reported outcome after
  breast reconstruction},'' \emph{\BIBforeignlanguage{en}{BMJ Open}}, vol.~11,
  no.~12, p. e052676, dec 2021.

\end{thebibliography}

\end{document}